\documentclass{article} %
\usepackage{iclr2027_conference,times}

\usepackage{amsmath,amsfonts,bm}

\def\eqref#1{equation~\ref{#1}}

\def\1{\bm{1}}

\DeclareMathAlphabet{\mathsfit}{\encodingdefault}{\sfdefault}{m}{sl}
\SetMathAlphabet{\mathsfit}{bold}{\encodingdefault}{\sfdefault}{bx}{n}

\usepackage[T1]{fontenc}   %

\usepackage[hidelinks]{hyperref}
\usepackage{url}
\usepackage{booktabs}
\usepackage{array}
\usepackage{graphicx}
\usepackage{amsmath}
\usepackage{amsthm}
\newtheorem{proposition}{Proposition}

\title{Finite-Sample Sensitivity of Kernel Metrics \\ for Generative Evaluation}

\author{Tomasz Radzikowski \& Mateusz Modrzejewski \\
Institute of Computer Science \\
Warsaw University of Technology \\
Warsaw, Poland}

\iclrfinalcopy
\begin{document}

\maketitle
\lhead{Preprint}

\begin{abstract}
Kernel distances such as the Kernel Inception Distance (KID) and the Kernel Audio Distance (KAD) compare generated and reference samples in an embedding space, and model comparisons rest on their values. Each ships with one fixed kernel, and a reported value does not say which changes that kernel detects at the sample size used. We run each default as a permutation two-sample test on real embeddings of symbolic music, audio and images, comparing a sample with an independent sample changed by a mean shift, an isotropic contraction, or a rotation that keeps the mean and the covariance trace and is matched to the shift in transport distance. At 780 samples per side, both defaults rarely detect the rotations in five evaluation spaces, while they detect the shifts in every run. The median-bandwidth Gaussian kernel of KAD detects the contraction and regains power against rotations on image embeddings with an order of magnitude more samples. The cubic kernel of KID contains second- and third-moment information but misses the contraction in most runs, and our experiments do not establish a high-power rotation regime for it over the tested sample sizes. Narrower Gaussian bandwidths restore sensitivity in music and audio, an $\ell_1$ Laplace kernel does so for images, and aggregated tests combine both at nominal level. We recommend that a reported kernel distance state its sample budget and the changes its configuration detects at that budget.
\end{abstract}

\section{Introduction}

Generative models of images, audio and music are compared by distances between embedded sample sets. The Fr\'echet family (FID, FAD, and FMD for symbolic music; \citealp{heusel2017gans,kilgour2018fr,fmd2025}) compares Gaussian fits of the embeddings. Kernel distances replace the Gaussian fit with a maximum mean discrepancy (MMD) under a fixed kernel: KID uses a cubic polynomial kernel \citep{binkowski2018demystifying}, KAD a Gaussian kernel whose bandwidth is set by the median heuristic \citep{chung2025kad}, and CMMD a Gaussian kernel on CLIP embeddings \citep{jayasumana2024rethinking}. Papers report these values to rank models and to argue that a change improved the quality or the diversity of samples.

Whether a reported value can reflect a given kind of difference depends on the kernel, on the geometry of the embedding, and on the number of samples. The two-sample testing literature establishes each dependence. The kernel determines which alternatives a test can detect \citep{gretton2012kernel,sutherland2017generative}, Gaussian-kernel power decays with dimension under common bandwidth choices \citep{ramdas2015decreasing}, and in high dimension MMD and energy-distance tests respond mainly to differences in marginal means and variances \citep{zhu2021interpoint,yan2023kernel}. These results do not tell a practitioner which differences the shipped KID or KAD configuration detects at the sample budget of an evaluation, because that depends on the embedding distributions in use and has to be measured on them.

We measure this on real embedding distributions of symbolic music (CLaMP~2), audio (VGGish and PANNs) and images (Inception features of CIFAR-10 and STL-10). We apply one known change to an independent sample, and test the two samples with each kernel configuration as a permutation test at level $0.05$. The changes are a mean shift, an isotropic contraction, a rotation that keeps the mean and the covariance trace, low-rank changes in the leading principal directions, and losses of coverage. We match strengths in 2-Wasserstein distance, so that the Gaussian transport bound of each rotation is at least the distance of the mean shift it is compared with, and we vary the sample size from $195$ to $20{,}000$ per side.

At $780$ samples per side, the median heuristic detects the weak rotation in at most $4$ of $20$ runs in each music and audio space, and the matched mean shift in every run (Figure~\ref{fig:power}). The cubic kernel behaves the same way on both. The two defaults separate on scale and on sample size. The median heuristic detects an isotropic contraction of matched transport distance in every run in all five spaces, while the cubic kernel misses it in most runs. On Inception features, the median heuristic rejects the rotation in every run from $10{,}000$ samples per side, while the cubic kernel rejects it in $2$ of $5$ runs at $20{,}000$ (Figure~\ref{fig:samplesize}).

Two expansions clarify which information the kernels retain. At wide bandwidths, the Gaussian MMD equals the squared mean difference divided by the squared bandwidth, and covariance differences enter only at the next order (Appendix~A.1). The cubic kernel's MMD is an exact sum of mean, second-moment and third-moment terms, with the second-moment term weighted by a further factor $1/d$ (Section~\ref{sec:mechanism}). The cubic kernel therefore contains covariance information, and its observed weakness concerns the finite-sample signal-to-noise ratio of the full statistic. The kernel that restores sensitivity depends on the embedding. A Gaussian bandwidth of $0.05$ to $0.1$ times the median squared distance suffices in music and audio. On image features, the only single kernel among those we tried that reaches high power is an exponential kernel on $\ell_1$ distances, consistent with the result of \citet{zhu2021interpoint} that $\ell_1$-distance tests respond to differences in marginal distributions where $\ell_2$-distance tests respond to means and variances.

This paper makes three contributions. First, we measure the finite-sample sensitivity of the default KID and KAD kernels to controlled changes of real evaluation embeddings in five spaces, so that a reported value can be read against the changes its configuration detects (Section~\ref{sec:defaults}). Second, we separate the effects of sample size, Gaussian bandwidth and kernel family, which shows why a configuration that succeeds in one space fails in another (Sections~\ref{sec:samplesize} and~\ref{sec:mechanism}). Third, we turn these results into a reporting protocol that separates distributional testing from the description of location, covariance and spectral diversity and from checks for copying (Section~\ref{sec:spectrum}).

\textbf{Scope.} Our experiments evaluate statistical sensitivity in the chosen embedding spaces. A change with larger transport distance need not be more perceptually important, and higher rejection power does not by itself establish a better model ranking. We therefore read the results as an audit of metric configurations and sample budgets rather than as a comparison of perceptual quality measures.

\section{Related work}
\label{sec:related}

\textbf{Evaluation metrics.} FID \citep{heusel2017gans} and FAD \citep{kilgour2018fr} compare Gaussian fits of embeddings. Their small-sample bias motivated unbiased kernel estimators \citep{binkowski2018demystifying,jayasumana2024rethinking}, most recently KAD \citep{chung2025kad}, which adopts the median heuristic. \citet{otad2026} reports that KAD depends on the PANNs variant and on how the median is computed, and that KAD stays near its noise floor under rank-one contamination of audio embeddings, an embedding-level change close to our low-rank alternatives. \citet{nac2025} find that FAD agrees better than MMD with listener ratings of codec degradations. These studies examine signal-level and embedding-level perturbations and agreement with listeners. We examine which distributional changes the default kernels detect at a given sample size. \citet{practicalguide2024} vary the sample size, the dimension and the Gaussian bandwidth for several distances on synthetic and scientific data, and \citet{xu2018empirical} and \citet{stein2023exposing} compare metrics against desiderata. \citet{obray2022evaluation} show that MMD-based evaluation of graph generative models depends on the kernel and its parameters. We apply that template to neural embeddings with constructed alternatives. Concurrently, \citet{chen2026what} runs FID and KID as permutation tests on Inception features of CIFAR-10, with sweeps of dimension and sample size, and proposes a diagnostic that combines six tests. That study also finds KID weak against changes of dispersion. Our study differs in its object (the shipped defaults, including KAD's median heuristic), its domains (symbolic music, audio and images) and its sample range (up to $20{,}000$ per side).

\textbf{Kernel two-sample testing.} The choice of kernel governs the power of a test \citep{gretton2012kernel,sutherland2017generative}. \citet{ramdas2015decreasing} show that Gaussian-kernel power decreases with dimension under common bandwidth choices. \citet{zhu2021interpoint} prove that in high dimension, with small or moderate samples, energy-distance and Gaussian or Laplace MMD tests mainly target differences in marginal means and variances, and that tests on $\ell_1$ distances capture differences in marginal distributions. \citet{yan2023kernel} derive which moment discrepancies a kernel test detects as dimension and sample size grow. Our measurements are finite-sample, in specific embedding spaces, and for the default configurations. MMDAgg \citep{schrab2023mmdagg} aggregates tests over a collection of kernels with non-asymptotic level control, and MMD-FUSE \citep{biggs2023mmdfuse} fuses kernels without data splitting. We evaluate MMDAgg as an existing remedy on evaluation embeddings and report its cost.

\textbf{Diversity diagnostics.} The Vendi score \citep{friedman2023vendi} measures diversity as the exponentiated entropy of a similarity-kernel spectrum. The spectral summaries of Section~\ref{sec:spectrum} apply the same functional to the embedding covariance spectrum, normalized by the reference corpus.

\section{Experimental protocol}
\label{sec:protocol}

\textbf{Embedding spaces.} We use CLaMP~2 \citep{wu2024clamp} embeddings (768 dimensions) of the folk-rnn v2 training corpus of $23{,}636$ transcriptions \citep{sturm2016music}; VGGish \citep{hershey2017cnn} (128 dimensions) and PANNs \citep{kong2020panns} (2048 dimensions) embeddings of an eight-genre audio corpus of $1{,}999$ clips derived from FMA \citep{defferrard2017fma}; and InceptionV3 features (2048 dimensions) of CIFAR-10 and of STL-10. These are the five evaluation spaces. The two image spaces share an encoder. A sixth, out-of-domain symbolic encoder (Adversarial-MidiBERT, \citealp{zhao2025let}) enters only the analysis of Section~\ref{sec:mechanism}.

\textbf{Alternatives.} Each test compares two disjoint samples $A$ and $B$ of $n$ points drawn from one corpus, with $B$ transformed by exactly one map $T$. With $\mu$ the mean of $B$ and $r=\sqrt{\operatorname{tr}\operatorname{cov}(B)}$ its radius, the maps are a \emph{mean shift} $B + s\,r\,v$ with $v$ a random unit vector; a \emph{rotation} $(B-\mu)M+\mu$ with $M$ an orthogonal Cayley transform whose skew part is scaled by $s/(2\sqrt d)$, which keeps the mean, the covariance trace and the covariance eigenvalues; and an isotropic \emph{contraction} $\mu+(1-s)(B-\mu)$. Section~\ref{sec:defaults} adds low-rank changes and losses of coverage. We report weak ($s{=}0.1$) and strong ($s{=}0.5$) strengths. Unless noted, $n{=}780$, the size of the smallest per-genre subset of the symbolic corpora.

\textbf{Matching by transport distance.} We compute 2-Wasserstein distances in the Gaussian approximation (the Bures formula) from sample means and covariances. This is exact for the mean shift and for the contraction, whose scaling map is the optimal transport map. For a rotation of a non-Gaussian cloud, the formula gives the lower bound of \citet{gelbrich1990formula}. Matching the bound for the rotation to the exact distance of the shift therefore makes the rotated distribution at least as far from the original in transport distance as the shifted one. At $s{=}0.1$ the bound for the rotation is $1.07$ to $1.37$ times the distance of the shift, and bisection to equality gives $s{=}0.073$ to $0.093$. Equal transport distance does not imply equal statistical difficulty or equal perceptual severity. We use it to fix the comparison.

\textbf{Configurations and tests.} The KAD default is the Gaussian kernel $\exp(-\|x-y\|^2/(2\sigma^2))$ with $\sigma$ the median pairwise distance. The KID default is the cubic kernel $(1+x^\top y/d)^3$. We also test Gaussian kernels with $\sigma^2$ a fixed multiple of the median squared distance, an exponential kernel on $\ell_1$ distances at the median scale, MMDAgg at the defaults of \texttt{mmdagg} 1.0.0, and the Fr\'echet distance with plug-in and shrinkage covariance estimates. Single-kernel MMD configurations use the unbiased MMD$^2$ estimator and $500$ label permutations at level $\alpha{=}0.05$. MMDAgg uses the aggregation and calibration procedure of \texttt{mmdagg} 1.0.0. The Fr\'echet distance and the precision, recall, density and coverage statistics use the permutation budgets given in Appendices~A.10 and A.16. Each cell uses $20$ independent draws of samples and transformation. Cells that support a claim and whose count falls between $3$ and $17$ are extended to $50$ draws. Because the extension depends on the first draws, intervals for extended cells are descriptive. Draws resample a fixed corpus, so every interval describes variation conditional on that corpus. Empirical level comes from untransformed controls. It is at most $3/20$ per cell in the main grids and $4/20$ in one cell of the dimension ladder of Appendix~A.6.

\textbf{Centering in the real-pair analysis.} The real-pair arms of Section~\ref{sec:spectrum} subtract each side's sample mean once, before the permutations. This forces the observed difference of means to zero, while permuted splits generally have nonzero differences of means. For the linear part of the MMD, the observed split therefore minimizes the statistic over equal-size splits (Appendix~A.3). We report these arms as descriptive diagnostics. Their rejection counts are not estimates of calibrated power for detecting differences beyond an unknown location shift.

\section{Location, scale and orientation at 780 samples per side}
\label{sec:defaults}

\begin{figure}[t]
\begin{center}
\includegraphics[width=\textwidth]{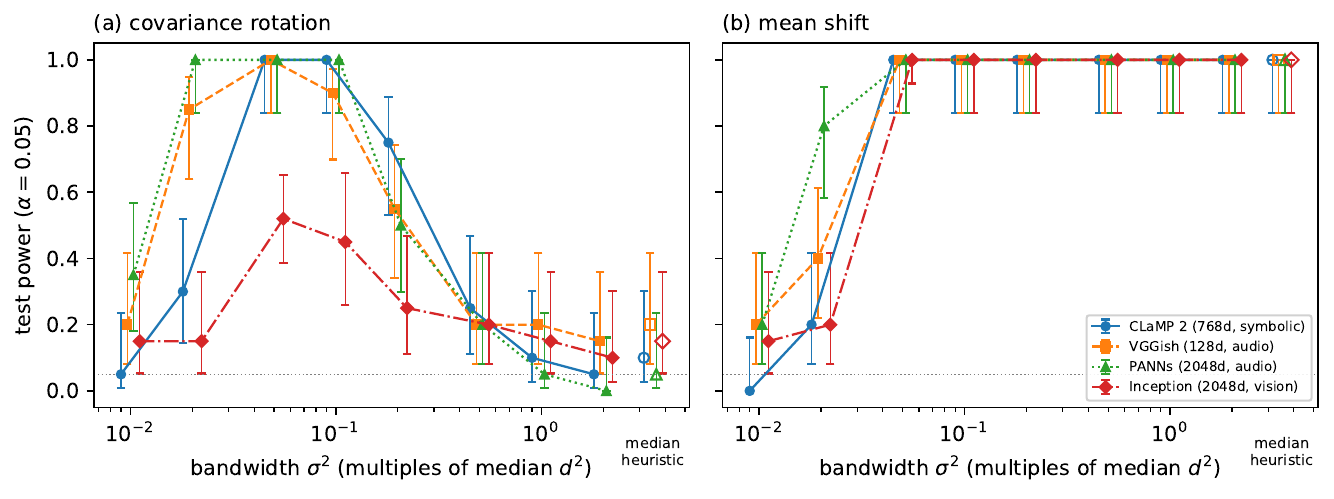}
\end{center}
\caption{Gaussian-kernel MMD rejection rates for a rotation that keeps the mean and the covariance trace (left) and for a mean shift (right), at $n{=}780$ per side and level $0.05$, against the bandwidth as a multiple of the median squared distance. Hollow markers show the median heuristic. Bars are descriptive Wilson $95\%$ intervals over $20$ draws, and over $50$ for the Inception point at $0.05$ ($26/50$). The shift has no greater transport distance than the rotation.}
\label{fig:power}
\end{figure}

\begin{table}[t]
\caption{Rejections at $n{=}780$ per side. Music and audio cells give the range over CLaMP~2, VGGish and PANNs. The window is the Gaussian kernel at $0.05\times$ the median squared distance. A dash marks a cell the battery did not run. Fr\'echet uses $n{=}390$ in the 2048-dimensional spaces (PANNs, Inception).}
\label{tab:main}
\begin{center}\small\setlength{\tabcolsep}{2.6pt}
\begin{tabular}{llccccc}
\toprule
Alternative & Spaces & median & cubic & window & MMDAgg & Fr\'echet \\
\midrule
mean shift, $s{=}0.1$ & music, audio & 20/20 & 20/20 & 20/20 & -- & 5--20/20 \\
 & Inception & 20/20 & 20/20 & 20/20 & 20/20 & 13/20 \\
contraction, $s{=}0.1$ & music, audio & 20/20 & 0--3/20 & -- & -- & 5--20/20 \\
 & Inception & 20/20 & 4/20 & 20/20 & -- & 13/20 \\
rotation, $s{=}0.1$ & music, audio & 1--4/20 & 0--2/20 & 20/20 & 18/20--50/50 & 10--20/20 \\
 & Inception & 3/20 & 3/20 & 26/50 & 20/20 & 16/20 \\
rank-2 rotation & CLaMP~2, PANNs & 4--7/50 & 0--2/50 & 30--39/50 & 36--44/50 & -- \\
 & Inception & 5/50 & 2/50 & 15/50 & 21/50 & -- \\
rank-1 inflation & CLaMP~2, PANNs & 40--50/50 & 2--8/50 & 43--50/50 & 50/50 & -- \\
 & Inception & 44/50 & 0/50 & 35/50 & 50/50 & -- \\
mode drop, mean-matched & music, audio & 1/20--4/50 & 0/20--3/50 & 2/20--34/50 & 0/20--50/50 & 1--9/20 \\
mode drop, least-displacing & all five & \multicolumn{4}{c}{0--7/50 in every configuration} & 1--4/20 \\
control & all five & \multicolumn{4}{c}{at most 3/20 per cell} & 0--2/20 \\
\bottomrule
\end{tabular}\end{center}
\end{table}

\subsection{The defaults detect mean shifts and rarely detect rotations}

At the median heuristic, the weak rotation is detected in at most $4$ of $20$ runs in every space of Figure~\ref{fig:power}, and the matched mean shift in every run (Table~\ref{tab:main}). The cubic kernel behaves the same way on both. Matching the rotation's transport bound exactly to the shift leaves the result unchanged (Appendix~A.2). In the units that the published implementations report, the equal-transport rotation moves KID and KAD by less than one standard deviation of a noise floor estimated from independent reference splits, and the mean shift moves them by nine to twenty-nine (Appendix~A.8). A practitioner reading either value at this sample size would see the shift and would not see the rotation.

A narrower Gaussian bandwidth detects the rotation in music and audio. At $0.05$ to $0.1$ times the median squared distance, the rotation is detected in $18$ to $20$ of $20$ runs in all three spaces. The useful range is two-sided, because at $0.01$ times the median the mean shift is detected in at most $4$ of $20$ runs (Figure~\ref{fig:power}).

\subsection{Contraction separates the two defaults}

At $s{=}0.1$, an isotropic contraction has the same transport distance as the mean shift of the same strength. The median heuristic detects it in every run in all five spaces, while the cubic kernel detects it in at most $6$ of $20$ (Table~\ref{tab:main}, Appendix~A.15). The Gaussian default therefore responds to location and to total scale, and the cubic kernel at this sample size responds mainly to location. The same difference appears for coverage: an intra-mode collapse that shrinks every point toward its cluster centroid is detected by every configuration except the cubic kernel (Appendix~A.11).

\subsection{Low-rank changes}

In seven model and reference pairs, the five leading eigendirections of the covariance difference hold $40$ to $87\%$ of its squared Frobenius norm (Appendix~A.17). We therefore apply mean-preserving changes inside the leading principal subspace: rotations within the top two and top five components, and inflation or deflation of the first component by $30\%$. The cubic kernel has low rejection rates for all four alternatives in the three spaces tested (Table~\ref{tab:main}). The median heuristic is substantially more sensitive to the rank-one rescalings than to the rotations. The window bandwidth and MMDAgg detect the rotations in music and audio, and on Inception features the rank-two rotation is hard for every configuration. In a CLaMP~2 control, the same inflation along a random direction and a displacement of $2\%$ of the points were detected in at most $2$ of $20$ runs by every configuration (Appendix~A.2). Changes along leading variance directions were easier to detect than these random-direction changes.

\subsection{Coverage loss}

Removing half of the largest of ten $k$-means modes moves the sample mean, and the defaults detect the loss through that displacement. When the displacement is subtracted (the mean-matched mode drop), the defaults miss the remaining difference (Table~\ref{tab:main}). What the window bandwidth and MMDAgg detect then depends on how the deleted points are replaced. Refilling them by resampling the remainder introduces duplicates in $8$ to $10\%$ of the sample, and duplicates alone, with no mode removed, are detected by both configurations in every run. Refilling from a disjoint reserve of the corpus removes that confound. Detection then ranges from $0$ of $20$ runs (MMDAgg, VGGish) to $50$ of $50$ (MMDAgg, PANNs) (Appendix~A.2). A least-displacing mode drop, which deletes half of the mode whose removal moves the mean least, has low rejection rates under every tested configuration, including MMDAgg and the Fr\'echet test (Appendix~A.11). Our conclusions about coverage are therefore local to these constructions. Appendix~A.11 reports their residual mean displacement.

\section{Sample size and kernel family}
\label{sec:samplesize}

\begin{figure}[t]
\begin{center}
\includegraphics[width=\textwidth]{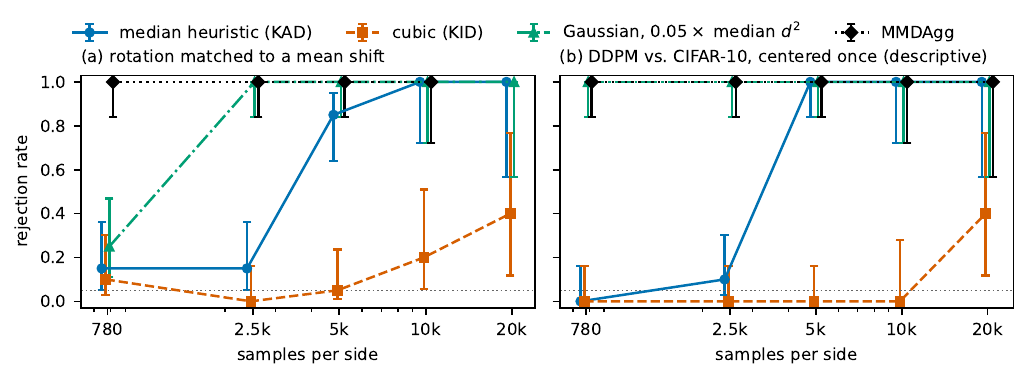}
\end{center}
\caption{Rejection rates against samples per side on Inception features of CIFAR-10, with descriptive Wilson $95\%$ intervals. Draws per point: $20$ up to $5{,}000$, $10$ at $10{,}000$, $5$ at $20{,}000$. (a) A rotation matched to a mean shift in transport distance. (b) DDPM samples against CIFAR-10 with each side centered once (Section~\ref{sec:protocol}). Panel (b) is descriptive: centering each side before permutation treats the observed and permuted labels differently, so its rates do not estimate calibrated power. MMDAgg was not run on (a) at $20{,}000$. The dotted line marks the nominal level $0.05$.}
\label{fig:samplesize}
\end{figure}

\subsection{The median heuristic regains power with more samples}

On Inception features of CIFAR-10, the median heuristic rejects the matched rotation in $3$ of $20$ runs at $780$ samples per side and in every run from $10{,}000$ (Figure~\ref{fig:samplesize}a). It is sensitive to this difference at a sample size that image datasets afford and that per-genre symbolic subsets do not. On synthetic Gaussians in 768 dimensions, the median heuristic likewise needs about an order of magnitude more samples than narrow bandwidths and MMDAgg to reach full power (Appendix~A.2). The cubic kernel rejects the rotation in $2$ of $10$ runs at $10{,}000$ and $2$ of $5$ at $20{,}000$. Five draws are too few to establish a trend at $20{,}000$, and we do not establish a high-power regime for the cubic kernel within the tested range. The window bandwidth and MMDAgg detect the rotation in every run from $2{,}500$ samples per side on.

\subsection{The kernel that restores power depends on the embedding}
\label{sec:mmdagg}

In music and audio, a fixed narrow Gaussian bandwidth is enough, and Gaussian-only aggregation detects the rotation in at least $84\%$ of runs in all three spaces. On Inception features of CIFAR-10, the best single Gaussian bandwidth reaches $26$ of $50$ runs, and Gaussian-only aggregation reaches $5$ of $20$ (Appendix~A.2). An exponential kernel on $\ell_1$ distances at the median scale, $\exp(-\|x-y\|_1/s)$, detects the rotation in $20$ of $20$ runs, where the same kernel on $\ell_2$ distances reaches $5$ of $20$ (Appendix~A.6). Varying smoothness and tail weight does not reproduce the gap: smoother and lighter-tailed kernels do better than rougher and heavier-tailed ones, and none of them reaches the count of the $\ell_1$ kernel (Appendix~A.6). Among the variations tested, these ablations favor the norm as the explanation. The $\ell_1$ advantage disappears on principal-component truncations of the same features, so in these experiments it is specific to the full-space representation. This agrees with the analysis of \citet{zhu2021interpoint}, in which $\ell_1$-distance tests detect differences in marginal distributions that $\ell_2$-distance tests miss in high dimension.

MMDAgg at package defaults combines Gaussian and Laplace kernels and detects the weak rotation in $18$ of $20$ to $50$ of $50$ runs in every space, at a pooled empirical level of $13/260$. Its value is that it does not require choosing the successful family in advance, at about $16$ times the wall-clock cost of a single kernel (Appendix~A.7). It does not detect every alternative, as the least-displacing mode drop shows.

\section{Why the defaults differ}
\label{sec:mechanism}

For the cubic kernel $k(x,y)=(1+x^\top y/d)^3$ and distributions with finite third moments, expanding the kernel gives the exact identity
\begin{equation}
\mathrm{MMD}^2_k(P,Q)=\frac{3}{d}\|\Delta\mu\|^2+\frac{3}{d^2}\|\Delta M_2\|_F^2+\frac{1}{d^3}\|\Delta M_3\|_F^2,
\label{eq:cubic}
\end{equation}
where $M_j=\mathbb{E}[X^{\otimes j}]$ is the raw moment tensor of order $j$. A covariance difference with equal means therefore gives a positive population KID. The same identity holds for the V-statistic with sample moments, and the unbiased estimator splits exactly into the three corresponding kernels $3u$, $3u^2$ and $u^3$ with $u=x^\top y/d$. The cubic kernel represents covariance. Its coefficients alone are not a power analysis, because moment magnitudes, the third-moment term, feature scale and permutation variability all enter. The observed weakness concerns the finite-sample signal-to-noise ratio of the full statistic.

For the Gaussian kernel, the leading term at wide bandwidth is $\|\Delta\mu\|^2/\sigma^2$, and covariance differences enter at order $\sigma^{-4}$ (Proposition~1, Appendix~A.1). At the median bandwidth, the ratio of the kernel statistic to $\|\Delta\mu\|^2/\sigma^2$ is $0.64$ to $0.67$ across fifteen pairs of symbolic corpora. This expansion describes a weighting that the median bandwidth can fall into. It does not by itself set the sample size at which the weighting stops mattering, which Section~\ref{sec:samplesize} measures.

Effective dimension is associated with the regime but does not determine it. Under Adversarial-MidiBERT, whose embeddings of the folk corpus have effective dimension $11.2$, the median heuristic detects the weak rotation in $20$ of $20$ runs. The music and audio spaces, of intermediate effective dimension, need a narrow bandwidth. In the image spaces, of the highest effective dimension, the best single Gaussian bandwidth reaches about half power or less (Appendix~A.6). Within one space the association breaks: a 16-component truncation of the Inception features has effective dimension $12.8$, and the median heuristic detects the rotation there in only $4$ of $20$ runs (Appendix~A.6).

\section{Implications for generative evaluation}
\label{sec:spectrum}
\label{sec:incumbent}

\textbf{Real model pairs.} We ran ten model and data comparisons, in twelve pair--encoder combinations, through the battery without centering (Appendix~A.12). Every configuration detects every pair, with one exception: the window bandwidth misses the Aria pair, which the defaults and MMDAgg detect. At these sample sizes, any configuration distinguishes these models from their data. The choice of configuration matters for smaller samples, for finer comparisons such as model against model, and for interpreting what differs.

\textbf{Descriptive effect of separate centering.} With each side centered once (Section~\ref{sec:protocol}), the cubic kernel rejects the GPT-2-class model against its training corpus in $0$ of $50$ runs, the median heuristic in $25$ of $50$, and the window bandwidth and MMDAgg in $50$ of $50$. The DDPM pair of Figure~\ref{fig:samplesize}b shows the same pattern, and the Fr\'echet test rejects the centered pairs in every run in all four spaces tested (Appendix~A.10). These counts describe the once-centered procedure. Because its preprocessing treats the observed and permuted labels differently, they do not establish comparative power for a calibrated test after removal of an estimated location difference (Appendix~A.3).

\textbf{The Fr\'echet distance as a test.} Run through the same permutation protocol, the Fr\'echet distance detects the weak rotation in every run in CLaMP~2 and VGGish. In the 2048-dimensional spaces we run it at $n{=}390$, where its rejection rate against the weak constructed changes is $0.25$ to $0.80$, so these cells are not a matched-budget comparison with the kernels (Appendix~A.10). No square root failed in $196{,}520$ evaluations. Like the default kernels, it rarely detects the mean-matched mode drop.

\textbf{A reporting protocol.} Table~\ref{tab:protocol} separates distributional testing from descriptions of location, covariance and spectral concentration, and from checks for copying of training data. Appendix~A.20 illustrates the spectral summaries on a temperature sweep of a symbolic transformer. A copying check needs its own reference set, statistic and calibration.

\begin{table}[t]
\caption{A reporting protocol. Each row is a separate question with its own quantity and calibration.}
\label{tab:protocol}
\begin{center}\small\setlength{\tabcolsep}{3pt}
\begin{tabular}{>{\raggedright\arraybackslash}p{2.5cm}>{\raggedright\arraybackslash}p{3.3cm}>{\raggedright\arraybackslash}p{3.6cm}>{\raggedright\arraybackslash}p{3.6cm}}
\toprule
Question & Quantity & Disclosure & Limitation \\
\midrule
Do the distributions differ? & a two-sample test chosen in advance (MMDAgg at package defaults) & kernel collection, preprocessing, samples per side, calibration, $p$-value & non-rejection is not equivalence; rejection does not measure quality \\
Do the locations differ? & mean displacement in embedding units & normalization, uncertainty & says nothing about spread or support \\
How do covariances differ? & Bures term on shrinkage covariances, or a normalized covariance difference & estimator, shrinkage, rank, uncertainty & hard to estimate when $n<d$ \\
Does spectral diversity differ? & effective rank and $n_{90}$ against the reference & matched sample sizes, reference definition & misses rotations and many coverage changes \\
Are training samples copied? & a dedicated copying test against held-out training data & reference and training sets, statistic, threshold, calibration & no detection is not proof of no memorization \\
\bottomrule
\end{tabular}\end{center}
\end{table}

\section{Limitations and conclusion}
\label{sec:discussion}

\textbf{Limitations.} We measure statistical sensitivity in embedding space, with constructed changes as controlled probes. They are not models of specific generator failures. The once-centered real-pair analyses are descriptive because their preprocessing treats the observed and permuted labels differently. They do not support our calibrated power claims. Draws resample fixed corpora, and two of the five evaluation spaces share an encoder. MMD-FUSE, learned kernels, classifier two-sample tests and CMMD in CLIP space were not run, because the package defaults are the object of study. Learned kernels would also need data splitting, which is costly at $n{=}780$. Perceptual validation is out of scope. Split-based $z$-scores change with the duplicate structure of a corpus (Appendix~A.21), so we state every claim in permutation power.

\textbf{Conclusion.} We measured the finite-sample sensitivity of the default kernels of KID and KAD on real embeddings of music, audio and images. At $780$ samples per side, both defaults rarely detect rotations that keep the mean and the covariance trace, although these rotations are matched in transport distance to mean shifts that both defaults detect in every run. The median-bandwidth Gaussian kernel also detects changes of scale and regains sensitivity to rotations with an order of magnitude more samples. The cubic kernel misses most contractions at $780$ samples and stays weak against rotations within the tested sample range, although its population discrepancy contains the covariance difference. The kernel that restores sensitivity depends on the embedding: a narrow Gaussian bandwidth for music and audio, and an $\ell_1$ Laplace kernel for image features. We recommend that every reported kernel distance state its sample budget and the changes its configuration detects at that budget.

\subsubsection*{AI use statement}

In this work, we used generative AI tools to implement methods (analysis code and experimental tooling) and to assist in the writing of derivations, both of which are tasks with required disclosure. We did not use generative AI tools to generate synthetic datasets, to propose or refine hypotheses, to design the methodology or experiments, or to interpret results. We also used generative AI tools to give feedback on drafts of the manuscript, including its statistical presentation, and the authors assessed every suggestion before adopting it. The remaining required-disclosure tasks (translation, dataset cleaning, qualitative data analysis) are not applicable to this work. Additionally, we used generative AI tools for tasks with recommended disclosure: drafting and editing parts of the manuscript, creating and editing software code, formatting references, and searching for related work. We have reviewed all AI-assisted work: every reference found with AI assistance was verified against its original source before citation, all experiments were executed on the authors' infrastructure from seeded scripts, AI-assisted code was tested against synthetic cases with known answers, and the derivations of Appendix~A.1 were verified numerically against closed-form expectations. We take responsibility for the final content of this work, including text, claims, and artifacts produced with the aid of generative AI.

\subsubsection*{Ethics statement}
This work studies evaluation methodology on public datasets (CIFAR-10, STL-10, an FMA-derived audio corpus, MidiCaps, Aria-MIDI, and the folk-rnn v2 corpus) and on generative models used under their published licenses, and involves no human subjects or personally identifiable data. We foresee no harmful applications beyond those of evaluation methodology generally. The copying analysis is intended to support training-data accountability rather than to circumvent it.

\subsubsection*{Reproducibility statement}
Every reported count maps to a per-cell result file that records the seed of each draw. The files and the scripts are available from the authors on request. Seeds derive from a stable checksum of the cell identifier, so a cell reproduces across processes. Sample survival across the temperature sweep is reported in Appendix~A.4. The symbolic corpus is the public folk-rnn v2 training set of \citet{sturm2016music}, available in the folk-rnn repository (\url{https://github.com/IraKorshunova/folk-rnn}). We converted it to one ABC file per tune, and the conversion script is available on request. Section~\ref{sec:protocol} specifies the alternatives and the testing protocol. Appendix~A.3 gives the exact constructions and the centering procedures. Appendices~A.2 and A.5 to A.21 tabulate the cells behind the figures and claims, and Appendix~A.1 contains the derivations and their numerical checks. MMDAgg is used at the package defaults of \texttt{mmdagg} 1.0.0 without per-comparison tuning. All analysis runs on CPU except embedding extraction.

\bibliography{iclr2027_conference}
\bibliographystyle{iclr2027_conference}

\appendix
\section{Appendix}

\textbf{A.1 Wide-bandwidth degeneracy of the Gaussian MMD.}
\begin{proposition}
Let $P,Q$ on $\mathbb{R}^d$ have means $\mu_P,\mu_Q$, covariances $\Sigma_P,\Sigma_Q$, and finite fourth moments, and let $k_\sigma(x,y)=\exp(-\|x-y\|^2/(2\sigma^2))$. Then as $\sigma^2\to\infty$,
\[\mathrm{MMD}^2_{k_\sigma}(P,Q) \;=\; \frac{\|\mu_P-\mu_Q\|^2}{\sigma^2} \;+\; O(\sigma^{-4}),\]
and covariance differences enter only at order $\sigma^{-4}$. If moreover $P,Q$ are Gaussian with $\mu_P=\mu_Q$, the leading term is
\[\mathrm{MMD}^2_{k_\sigma}(P,Q) \;=\; \frac{(\operatorname{tr}\Sigma_P-\operatorname{tr}\Sigma_Q)^2 \,+\, 2\,\|\Sigma_P-\Sigma_Q\|_F^2}{4\sigma^4} \;+\; O(\sigma^{-6}).\]
\end{proposition}
\emph{Derivation.} Write $\mathrm{MMD}^2 = \mathbb{E}k(x,x') + \mathbb{E}k(y,y') - 2\mathbb{E}k(x,y)$ and expand $k_\sigma = 1 - \tfrac{\|x-y\|^2}{2\sigma^2} + \tfrac{\|x-y\|^4}{8\sigma^4} - \cdots$. The order-$\sigma^0$ terms cancel. At order $\sigma^{-2}$, using $\mathbb{E}\|x-x'\|^2 = 2\operatorname{tr}\Sigma_P$, $\mathbb{E}\|y-y'\|^2 = 2\operatorname{tr}\Sigma_Q$, and $\mathbb{E}\|x-y\|^2 = \|\Delta\mu\|^2 + \operatorname{tr}\Sigma_P + \operatorname{tr}\Sigma_Q$, the trace terms cancel exactly and the coefficient is $\|\Delta\mu\|^2$. For the Gaussian equal-means case, $x-x'\sim\mathcal N(0,2\Sigma_P)$, $x-y\sim\mathcal N(0,\Sigma_P+\Sigma_Q)$, and the quartic identity $\mathbb{E}(z^\top z)^2 = (\operatorname{tr}S)^2 + 2\operatorname{tr}S^2$ for $z\sim\mathcal N(0,S)$ gives, after collecting the three terms at order $\sigma^{-4}$, the stated coefficient $2(\Delta\operatorname{tr}\Sigma)^2 + 4\|\Delta\Sigma\|_F^2$ over $8\sigma^4$. Empirically, the ratio of the kernel statistic to the first-order prediction $\|\Delta\mu\|^2/\sigma^2$ is constant across fifteen symbolic corpus pairs at the median bandwidth ($0.64$--$0.67$, s.d.\ $0.010$). The univariate wide-bandwidth limit is stated as a lemma by \citet{jacobs2026robust}, the derivation above being its multivariate counterpart. The bandwidth-dependence of kernel test power is established broadly by \citet{ramdas2015decreasing}. Both orders of the proposition were verified against closed-form Gaussian expectations (agreement to $6\times10^{-5}$ at $\sigma^2{=}10^6$). The cubic kernel has an exact counterpart, equation~(\ref{eq:cubic}): expanding $(1+u)^3$ with $u=x^\top y/d$ and using $(x^\top y)^j=\langle x^{\otimes j},y^{\otimes j}\rangle$ gives the mean, second-moment and third-moment terms with weights $3/d$, $3/d^2$ and $1/d^3$. The identity involves no limit and holds at every scale of the features, including VGGish, where $\mathbb{E}|u|$ is of order $10^4$ (Appendix~A.17). We checked it on sample moments to machine precision.

\textbf{A.2 Full power tables.} Table~\ref{tab:a2power} reports per-cell rejection counts behind Figure~\ref{fig:power} ($n{=}780$ per side, $20$ draws, $500$ permutations, $\alpha{=}0.05$). Table~\ref{tab:a2other} reports the fixed-kernel arms and aggregation, and Table~\ref{tab:a2synth} the synthetic sample-size analysis. Recalibrating the rotation by bisection so that its Gaussian $W_2$ bound equals the exact $W_2$ of the mean shift ($s{=}0.073$ to $0.093$) gives median-heuristic rejections of $1/20$, $4/20$ and $0/20$ in CLaMP~2, VGGish and PANNs, at control levels of $0$ to $1/20$, against $20/20$ for the mean shift.

\begin{table}[h]
\caption{Cells give rejections out of $20$ draws by Gaussian bandwidth, in multiples of the median $d^2$; ``med'' is the classical median heuristic $\sigma{=}\mathrm{med}(d)$. C2 is CLaMP~2, VG is VGGish, and PN is PANNs. The main text quotes the med column of this table.}
\label{tab:a2power}
\begin{center}\small\setlength{\tabcolsep}{4.5pt}
\begin{tabular}{llccccccccc}
\toprule
Distortion & Space & $.01$ & $.02$ & $.05$ & $.1$ & $.2$ & $.5$ & $1$ & $2$ & med \\
\midrule
rotation $s{=}.1$ & C2 & 1 & 6 & 20 & 20 & 15 & 5 & 2 & 1 & 2 \\
 & VG & 4 & 17 & 20 & 18 & 11 & 4 & 4 & 3 & 4 \\
 & PN & 7 & 20 & 20 & 20 & 10 & 4 & 1 & 0 & 1 \\
\midrule
mean shift $s{=}.1$ & C2 & 0 & 4 & 20 & 20 & 20 & 20 & 20 & 20 & 20 \\
 & VG & 4 & 8 & 20 & 20 & 20 & 20 & 20 & 20 & 20 \\
 & PN & 4 & 16 & 20 & 20 & 20 & 20 & 20 & 20 & 20 \\
\midrule
control ($s{=}0$) & C2 & 1 & 2 & 0 & 0 & 0 & 0 & 0 & 0 & 0 \\
 & VG & 1 & 0 & 1 & 1 & 0 & 0 & 0 & 0 & 0 \\
 & PN & 1 & 1 & 0 & 1 & 1 & 1 & 1 & 1 & 1 \\
\bottomrule
\end{tabular}\end{center}
\end{table}

\begin{table}[h]
\caption{Fixed-kernel arms and aggregation at weak strength $s{=}0.1$ unless noted, with $20$ draws per cell and $50$ where two counts appear or where marked. Median-baseline rotation cells with two counts report seeds $0$;$1$. $\dagger$: a second run of this cell; the real-comparisons battery gives $25/50$ for the same configuration (\S\ref{sec:spectrum}).}
\label{tab:a2other}
\begin{center}\small
\begin{tabular}{lccc}
\toprule
 & CLaMP~2 & VGGish & PANNs \\
\midrule
cubic (KID): rotation & 1/20 & 2/20 & 0/20 \\
cubic: contraction & 4/20 & 2/20 & 2/20 \\
cubic: mean shift & 20/20 & 20/20 & 20/20 \\
cubic: control & 0/20 & 0/20 & 1/20 \\
\midrule
MMDAgg (defaults): rotation & 18/20 & 50/50 & 20/20 \\
MMDAgg, Gaussian-only: rotation & 18/20 & 42/50 & 20/20 \\
MMDAgg: rotation $s{=}0.5$ & 20/20 & 20/20 & 20/20 \\
MMDAgg: control (pooled, all five spaces) & \multicolumn{3}{c}{$13/260$} \\
median baseline: rotation & 2/20 & 3/20;\,4/20 & 1/20;\,2/20 \\
median baseline: control & 0/20 & 0/20 & 0/20 \\
\midrule
Inception (vision): MMDAgg / Gauss-only & \multicolumn{3}{c}{$20/20$ \;/\; $5/20$} \\
Inception: median / cubic: rotation & \multicolumn{3}{c}{$3/20$ \;/\; $3/20$} \\
Inception: best single bandwidth ($0.05$) & \multicolumn{3}{c}{$26/50$} \\
Inception: MMDAgg rotation $s{=}0.5$ & \multicolumn{3}{c}{$20/20$} \\
\midrule
recentered GPT-2 pair, median: $n{=}195/390$ & \multicolumn{3}{c}{$0/20$ \;/\; $0/20$} \\ \quad $n{=}780/1560$ & \multicolumn{3}{c}{$28/50^{\dagger}$ \;/\; $20/20$} \\
\quad window and MMDAgg, all $n$ & \multicolumn{3}{c}{$19$--$20/20$} \\
\midrule
STL-10: MMDAgg / Gauss-only & \multicolumn{3}{c}{$20/20$ \;/\; $7/20$} \\
STL-10: median / cubic: rotation & \multicolumn{3}{c}{$2/20$ \;/\; $2/20$} \\
STL-10: best single bandwidth ($0.05$) & \multicolumn{3}{c}{$8/20$ (mean shift $17/20$)} \\
\bottomrule
\end{tabular}\end{center}
\end{table}

\begin{table}[h]
\caption{Synthetic power on 768d Gaussians with equal means and $25\%$ of dimensions rescaled by $1{+}\gamma$, with $50$ repetitions per cell and $20$ for MMDAgg. The empirical level of the ``$1$'' arm at $\gamma{=}0$ is $0.06$. Headers are multiples of the median \emph{squared} distance, so column ``$1$'' ($\sigma^2{=}\mathrm{med}\,d^2$) is the synthetic proxy for the classical heuristic, which the real-data tables report separately as $\sigma{=}\mathrm{med}(d)$.}
\label{tab:a2synth}
\begin{center}\small\setlength{\tabcolsep}{5pt}
\begin{tabular}{llccccccc}
\toprule
$\gamma$ & $n$ & $.02$ & $.05$ & $.1$ & $.5$ & $1$ & $2$ & MMDAgg \\
\midrule
$0.05$ & $500$ & 1.00 & 1.00 & .92 & .24 & .10 & .06 & .95 \\
 & $2000$ & 1.00 & 1.00 & 1.00 & .96 & .46 & .24 & 1.00 \\
 & $6000$ & 1.00 & 1.00 & 1.00 & 1.00 & 1.00 & .72 & 1.00 \\
$0.15$ & $500$ & 1.00 & 1.00 & 1.00 & 1.00 & .96 & .66 & 1.00 \\
 & $2000$ & 1.00 & 1.00 & 1.00 & 1.00 & 1.00 & 1.00 & 1.00 \\
 & $6000$ & 1.00 & 1.00 & 1.00 & 1.00 & 1.00 & 1.00 & 1.00 \\
\bottomrule
\end{tabular}\end{center}
\end{table}

Pooled empirical level across all fixed-bandwidth Gaussian control cells (the three grids of Table~\ref{tab:a2power}, the Inception grid, and the STL-10 arms) is $35/760$ ($4.6\%$). The per-cell maximum is $3/20$, reached only in the Inception grid.

\begin{table}[h]
\caption{Mode drop battery with $20$ draws, $50$ where a count fell between $3$ and $17$; half of the largest of $k{=}10$ $k$-means clusters is removed. ``clean'' replaces deleted points from a disjoint reserve of the corpus; ``resampled'' replaces them by drawing from the remainder, which injects duplicates at the measured rate $f$; ``duplicates only'' applies that rate with no mode removed. ``mean-matched'' re-subtracts the induced mean displacement.}
\label{tab:a2mode}
\begin{center}\small\setlength{\tabcolsep}{4pt}
\begin{tabular}{llcccc}
\toprule
Space & Variant & median & cubic & window & MMDAgg \\
\midrule
CLaMP~2 & control & 1/20 & 1/20 & 2/20 & 1/20 \\
($f{=}0.10$) & clean, mean-matched & 4/50 & 2/50 & 34/50 & 24/50 \\
 & resampled, mean-matched & 15/50 & 4/50 & 50/50 & 50/50 \\
 & duplicates only & 6/50 & 4/50 & 50/50 & 50/50 \\
\midrule
VGGish & control & 1/20 & 1/20 & 1/20 & 1/20 \\
($f{=}0.084$) & clean, mean-matched & 1/20 & 0/20 & 2/20 & 0/20 \\
 & resampled, mean-matched & 7/50 & 2/50 & 50/50 & 50/50 \\
 & duplicates only & 0/20 & 1/20 & 20/20 & 20/20 \\
\midrule
PANNs & control & 1/20 & 0/20 & 2/20 & 1/20 \\
($f{=}0.094$) & & & & & \\
 & clean, mean-matched & 4/50 & 3/50 & 15/50 & 50/50 \\
 & resampled, mean-matched & 1/20 & 1/20 & 20/20 & 20/20 \\
 & duplicates only & 1/20 & 1/20 & 20/20 & 20/20 \\
\bottomrule
\end{tabular}\end{center}
\end{table}

\begin{table}[h]
\caption{Low-rank alternatives at $\gamma{=}0.3$, with $50$ draws where a count fell between $3$ and $17$ and $20$ otherwise; each construction keeps the mean of the transformed sample ($\|\Delta\hat\mu\|/r = 0.000$ between that sample before and after the transformation). Rank-2 and rank-5 are rotations confined to the leading principal components; inflation and deflation are rank-one scalings of the first component; ``random direction'' applies the same scaling along a random direction; contamination displaces $2\%$ of points.}
\label{tab:a2lowrank}
\begin{center}\small\setlength{\tabcolsep}{5pt}
\begin{tabular}{llcccc}
\toprule
Space & Alternative & median & cubic & window & MMDAgg \\
\midrule
CLaMP~2 & control & 1/20 & 0/20 & 2/20 & 1/20 \\
 & rank-2 rotation & 7/50 & 2/50 & 39/50 & 36/50 \\
 & rank-5 rotation & 16/50 & 3/50 & 50/50 & 50/50 \\
 & rank-1 inflation & 50/50 & 8/50 & 50/50 & 50/50 \\
 & rank-1 deflation & 50/50 & 5/50 & 50/50 & 50/50 \\
 & random direction & 2/20 & 1/20 & 1/20 & 1/20 \\
 & contamination $2\%$ & 2/20 & 0/20 & 0/20 & 0/20 \\
\midrule
PANNs & rank-2 rotation & 4/50 & 0/50 & 30/50 & 44/50 \\
 & rank-5 rotation & 17/50 & 1/50 & 50/50 & 50/50 \\
 & rank-1 inflation & 40/50 & 2/50 & 43/50 & 50/50 \\
 & rank-1 deflation & 36/50 & 5/50 & 50/50 & 50/50 \\
\midrule
Inception & rank-2 rotation & 5/50 & 2/50 & 15/50 & 21/50 \\
(CIFAR-10) & rank-5 rotation & 8/50 & 3/50 & 44/50 & 50/50 \\
 & rank-1 inflation & 44/50 & 0/50 & 35/50 & 50/50 \\
 & rank-1 deflation & 33/50 & 5/50 & 36/50 & 50/50 \\
\bottomrule
\end{tabular}\end{center}
\end{table}

\textbf{A.3 Distortion battery conventions and toolkit replication.} Perturbation seeds derive deterministically from (distortion index, strength index, draw index, global seed). The Cayley rotation uses $M=(I+S)^{-1}(I-S)$ with $S = s\,(A-A^\top)/(2\sqrt d)$, $A_{ij}\sim\mathcal N(0,1)$, so strengths are not numerically comparable across implementations that scale the skew part differently (the evaluation toolkit's convention is ${\sim}2\times$ stronger at equal $s$; all cross-tool comparisons in this paper are of qualitative profiles, never of raw values). Replicating the copying experiment with the toolkit's own estimator on the four symbolic and audio embedding sets confirms the profile of Table~\ref{tab:copying}: at $5\%$ injected copies, the narrow-bandwidth statistic responds at $|z|=3.5$--$4.7$ against $0.4$--$0.8$ for the median bandwidth and $\le1.5$ for the Fr\'echet distance, with a per-bandwidth standardized multi-bandwidth variant the most sensitive at $|z|=8.5$--$13.6$. On a corpus stored in genre order, taking the first half as the reference makes every configuration reject in $20/20$ runs, where random splits stay at $2$--$5$ of $50$. Full grids over all six distortion types (including mode drop) at three strengths, for the original and deduplicated symbolic corpus and both audio spaces, are in the per-cell result files.

In the centered real-pair arms, each side is centered on its own mean once, before the permutation loop. The observed split then has zero mean difference, while permuted splits of the pooled centered points keep an ordinary sampling difference of means. A permutation test is exact when the distribution of its statistic is invariant under the permutations under the null \citep{hemerik2018exact}, and this preprocessing treats the observed labels differently from the permuted ones. For the linear kernel the effect is exact: for a pooled sample with mean zero and total squared norm $C$, every equal split satisfies $\widehat{\mathrm{MMD}}^2_{\mathrm{lin},u}=\frac{2n-1}{2(n-1)}\|\bar x-\bar y\|^2-\frac{C}{n(n-1)}$, so the observed split minimizes the statistic over all splits. The cubic kernel contains this linear term with weight $3/d$, and separate centering also changes its raw second and third moments. We do not derive the effect for every nonlinear statistic. Same-population splits centered the same way say nothing about calibration, because the procedure is conservative on them by construction. A test after a translation estimated on independent data would test equality after that fixed translation. We leave that test to future work.

\textbf{A.4 Sample survival in the temperature sweep.} The feature banks contain $7{,}379$, $10{,}205$, $10{,}139$, $10{,}000$ and $9{,}353$ embedded outputs at temperatures $0.5$, $0.7$, $0.85$, $1.0$ and $1.2$ ($9{,}357$ outputs survived generation at $T{=}1.2$, and $4$ were lost in feature extraction). Table~\ref{tab:e3} uses $5{,}000$-output subsamples and describes the successfully generated, parsed and embedded outputs. These summaries do not estimate generation or parsing failure rates. Losses differ by direction: at low temperature the model enters repetition loops that never emit the end token (cut by a $1{,}500$-token guard set below the model's ${\sim}1{,}955$-token context) and produces sequences too degenerate for the notation converter, and at high temperature losses come from format drift. Low-temperature losses remove the most degenerate samples before embedding, so the ratios of Table~\ref{tab:e3} describe the surviving, parseable outputs and omit some severe failures.

\textbf{A.5 Observational corpus-pair analysis.} Complementing the interventional battery, we measure which term of the Fr\'echet decomposition $\|\Delta\mu\|^2 + \operatorname{tr}(\Sigma_r+\Sigma_t-2(\Sigma_r\Sigma_t)^{1/2})$ the kernel statistic follows on \emph{real} corpus pairs: fifteen pairs of genre subsets of a symbolic corpus \citep{melechovsky2024midicaps}, five independent subsamples of $6{,}000$ pieces, fourteen bandwidths, and two encoder versions (CLaMP~2/3, \citealp{wu2024clamp,wu2025clamp3}). Because the location and dispersion terms are collinear across pairs (Pearson $0.84$/$0.86$), the regression-based dispersion weight is reported with bootstrap intervals over pairs and alongside an apportionment-free statistic: the rank correlation between a pair's location share and its kernel-statistic residual from a linear fit on the Fr\'echet distance. Three statements survive the uncertainty analysis: below $\sigma^2{=}18$ the dispersion weight is $58$--$66\%$ across both encoders; above $\sigma^2{=}48$ it is $0.5$--$8.8\%$ (at the median bandwidth $\sigma^2{\approx}600$: $1.9\%$ and $1.6\%$; cubic kernel: $12.1\%$); and the residual correlation changes sign at $\sigma^2{=}32$ and $30$ for the two encoders, the transition whose location coincides with the power window of Figure~\ref{fig:power}. No claim about a peak survives the bootstrap (interval widths up to $84$ percentage points, driven by which corpora are compared rather than by sampling within them), which is why the claim rests on the sign change rather than on a maximum. The residual-correlation transition is present in symbolic pairs and absent in audio pairs. Deduplication and PCA preprocessing do not explain the difference, and it remains unexplained. The same sweep shows that a ranking can depend on the bandwidth: the pair second most distant at $\sigma^2{=}30$ is fifth at the median.

\textbf{A.6 Within-space effective-dimension manipulation.} Table~\ref{tab:a6} reports rotation power ($s{=}0.1$, $20$ draws) on PCA truncations and random orthonormal projections of Inception/CIFAR-10 embeddings.

\begin{table}[h]
\caption{Rotation power along the within-space $d_{\mathrm{eff}}$ ladder on Inception/CIFAR-10 at $s{=}0.1$, with $20$ draws unless marked. Columns $.02$--$.1$ give the Gaussian bandwidth as a multiple of the median $d^2$; ``shift $.05$'' is the mean-shift arm at the $0.05\times$ bandwidth; ``ctrl'' is the maximum control-cell count across the row's arms.}
\label{tab:a6}
\begin{center}\small\setlength{\tabcolsep}{5pt}
\begin{tabular}{lrccccccc}
\toprule
proj. & $d_{\mathrm{eff}}$ & $.02$ & $.05$ & $.1$ & median & MMDAgg & shift $.05$ & ctrl \\
\midrule
PCA-16 & $12.8$ & 18 & 16 & 11 & 4 & 18 & 19 & $\le3$ \\
PCA-32 & $23.0$ & 18 & 17 & 14 & 3 & 15 & 20 & $\le3$ \\
PCA-64 & $40.2$ & 13 & 16 & 13 & 3 & 12 & 20 & $\le3$ \\
PCA-128 & $69.0$ & 7 & 16 & 11 & 3 & 9 & 20 & $\le3$ \\
PCA-256 & $113.1$ & 4 & 13 & 12 & 3 & 10 & 20 & $\le3$ \\
PCA-512 & $167.9$ & 4 & 13 & 10 & 3 & 15 & 20 & $\le3$ \\
RP-64 & $46.2$ & 3 & 9 & 10 & 3 & 6 & 20 & $\le3$ \\
RP-256 & $125.9$ & 2 & 9 & 9 & 2 & 6 & 20 & $\le4$ \\
full & $266.3$ & 3 & $26/50$ & 9 & 3 & 20 & 20 & $\le3$ \\
\bottomrule
\end{tabular}\end{center}
\end{table}
The full-space $d_{\mathrm{eff}}{=}266.3$ here is the $50{,}000$-sample estimate. The protocol's $5{,}000$-sample subsets give $251.8$, the value used for comparison across spaces. Vocabulary coverage of the out-of-domain encoder of Section~\ref{sec:mechanism} was not a confound (rejects: $6/23{,}636$ and $1/10{,}000$ files). PANNs repeats the pattern (PCA-32/128: window $19$--$20/20$, median $1/20$). Aggregation is non-monotone along the ladder ($18\to9\to20$): the aggregate tracks the best member of its collection, whose Gaussian bandwidths weaken mid-ladder before the Laplace family, ineffective in the truncated subspaces but decisive in the full space (\S\ref{sec:mmdagg}), takes over. A sweep over kernel families narrows the explanation. Holding the functional form fixed and changing only the norm separates the kernels: at the median scale in the full Inception space, $\exp(-\|x-y\|_1/s)$ rejects the rotation in $20/20$ runs where $\exp(-\|x-y\|_2/s)$ reaches $5/20$ and the Gaussian $3/20$. On PCA truncations of the same embeddings the advantage disappears ($2/20$ at $k{=}64$, $5/20$ at $k{=}256$). Smoothness contributes little and in the direction opposite to the intuition that a peaked kernel helps: Mat\'ern kernels with $\nu = 1/2, 3/2, 5/2$ at the narrow scale give $7$, $9$ and $11$ of $20$. Tail weight also runs the wrong way: rational quadratic kernels with heavy tails ($\alpha{=}0.2$) reach $4$--$5/20$ against $10$--$16/20$ for light tails ($\alpha{=}5$). A first-order profile of kernel values does not track the power gap either. These ablations favor the norm over the tested smoothness and tail-weight variations as the explanation of the Laplace family's power in this space. The full-row $0.02$ cell is taken from the main vision grid ($20$ draws). The mean-shift arm at $0.05\times$ stays at $19$--$20/20$ along the ladder and every control cell stays within $0$--$4/20$, so rising rotation power at low $d_{\mathrm{eff}}$ reflects power rather than miscalibration. The $0.02\times$ mean-shift arm declines from $17/20$ to $5/20$ as $k$ grows (per-cell counts in the result files), repeating the two-sidedness of the window along the ladder.

\textbf{A.7 Cost.} The table reports seconds per test on one CPU with a fixed thread count, embedding extraction excluded, on Inception embeddings ($d{=}2048$) with $500$ permutations.

\begin{center}\small
\begin{tabular}{lrrrr}
\toprule
$n$ per side & median & cubic & window & MMDAgg \\
\midrule
$780$ & 0.8 & 0.6 & 0.6 & 10 \\
$5{,}000$ & 23 & 20 & 23 & 359 \\
$10{,}000$ & 90 & 76 & 89 & 1414 \\
\bottomrule
\end{tabular}\end{center}

\textbf{A.8 Effect sizes in practitioner units.} The table reports the displacement of the reported metric value, in standard deviations of a noise floor estimated from $20$ independent splits of the reference ($n{=}780$ per side, $5$ draws per distortion). KID is computed with \texttt{torchmetrics} and KAD with the authors' \texttt{kadtk}, both at package defaults.

\begin{center}\small\setlength{\tabcolsep}{5pt}
\begin{tabular}{lcccccccc}
\toprule
 & \multicolumn{2}{c}{CLaMP~2} & \multicolumn{2}{c}{VGGish} & \multicolumn{2}{c}{PANNs} & \multicolumn{2}{c}{Inception} \\
\cmidrule(lr){2-3}\cmidrule(lr){4-5}\cmidrule(lr){6-7}\cmidrule(lr){8-9}
Distortion & KID & KAD & KID & KAD & KID & KAD & KID & KAD \\
\midrule
rotation, equal transport & $-0.2$ & $-0.2$ & $0.1$ & $0.3$ & $-0.3$ & $0.6$ & $0.8$ & $0.6$ \\
mean shift, $s{=}0.1$ & $8.8$ & $8.6$ & $14.3$ & $17.2$ & $17.5$ & $18.9$ & $24.8$ & $28.8$ \\
mean-matched mode drop & $0.4$ & $0.3$ & $-0.6$ & $-0.5$ & $-0.1$ & $0.2$ & $-0.6$ & $-0.4$ \\
isotropic contraction & $0.5$ & $6.1$ & $-0.3$ & $11.1$ & $0.4$ & $12.8$ & $1.1$ & $20.3$ \\
real model pair & $48$ & $49$ & $128$ & $151$ & $588$ & $744$ & $71$ & $78$ \\
\bottomrule
\end{tabular}\end{center}
The real-pair row sets the scale: an equal-transport shape difference is two to three orders of magnitude smaller in reported units than the difference between a model and its data, while a mean shift of equal or smaller transport magnitude is within one to two orders.

\textbf{A.9 Real pairs through the full battery.} Three pairs were selected by a geometry screen over twelve candidates as the three with the lowest location share of the Fr\'echet decomposition ($0.198$, $0.244$, $0.253$), and the screen is in the per-cell result files. The table gives rejections out of $20$ draws at $n{=}780$ per side, with the vision pair also run at larger samples.

\begin{center}\small\setlength{\tabcolsep}{5pt}
\begin{tabular}{llcccc}
\toprule
Pair & Arm & median & cubic & window & MMDAgg \\
\midrule
DDPM vs.\ CIFAR-10 & plain & 20/20 & 20/20 & 20/20 & 20/20 \\
$n{=}780$ & recentered & 0/20 & 0/20 & 20/20 & 20/20 \\
 & split controls & 3/50 & 3/50 & 4/50 & 5/50 \\
$n{=}2500$ & recentered & 2/20 & 0/20 & 20/20 & 20/20 \\
$n{=}5000$ & recentered & 20/20 & 0/20 & 20/20 & 20/20 \\
\midrule
Stable Audio vs.\ FMA & plain & 20/20 & 20/20 & 20/20 & 20/20 \\
(PANNs) & recentered & 20/20 & 0/20 & 20/20 & 20/20 \\
 & split controls & 2/20 & 1/20 & 0/20 & 1/20 \\
\midrule
Aria vs.\ its own & plain & 20/20 & 20/20 & 0/20 & 20/20 \\
continuations & recentered & 0/20 & 0/20 & 0/20 & 0/20 \\
 & split controls & 1/20 & 1/20 & 1/20 & 1/20 \\
\bottomrule
\end{tabular}\end{center}

In the centered DDPM comparisons of this table, through $n{=}5{,}000$, the cubic configuration records no rejections and the median configuration reaches $20/20$ at $n{=}5{,}000$. Appendix~A.18 extends this analysis: the cubic configuration records $2/5$ rejections at $n{=}20{,}000$. These are descriptive counts under once-centering. The centered Aria pair is not rejected under any configuration. Both sides of that pair share a $512$-token prompt within a $2{,}048$-token sequence.

\textbf{A.10 The Fr\'echet distance through the battery.} We run the Fr\'echet distance as a permutation test. At $d\le768$ we use $n{=}780$, $500$ permutations, and both estimators, which agree in every cell. At $d{=}2048$ we use $n{=}390$ and $100$ permutations. The table reports the plug-in estimator, and where the shrinkage estimator also completed it agrees with the plug-in column to within three draws (Inception: rotation $14/20$, mean shift $14/20$, contraction $16/20$, mean-matched drop $5/20$; PANNs: radial truncation $20/20$, intra-mode collapse $20/20$). On the recentered real pair both estimators give $20/20$ in Inception and in PANNs.

\begin{center}\small\setlength{\tabcolsep}{4pt}
\begin{tabular}{lccccccc}
\toprule
 & \multicolumn{4}{c}{constructed, $s{=}0.1$} & \multicolumn{2}{c}{coverage} & real \\
\cmidrule(lr){2-5}\cmidrule(lr){6-7}\cmidrule(lr){8-8}
Space & control & rotation & mean shift & contraction & mean-matched & collapse & recentered \\
\midrule
CLaMP~2 ($d{=}768$) & 0--1/20 & 20/20 & 20/20 & 20/20 & 9/20 & 20/20 & 20/20 \\
VGGish ($d{=}128$) & 0/20 & 20/20 & 18/20 & 17/20 & 1/20 & 20/20 & 20/20 \\
Inception ($d{=}2048$) & 2/20 & 16/20 & 13/20 & 13/20 & 4/20 & 20/20 & 20/20 \\
PANNs ($d{=}2048$) & 0/20 & 10/20 & 5/20 & 5/20 & 2/20 & 20/20 & 20/20 \\
\bottomrule
\end{tabular}\end{center}
Least-displacing mode drop is $1$--$4/20$ in every space, and radial truncation is $20/20$ except in Inception, where it is $12/20$. At $n{=}780$, the three weak constructed alternatives are rejected in $17$--$20/20$ runs in VGGish and $20/20$ in CLaMP~2. At $n{=}390$, their counts are $5$--$10/20$ in PANNs and $13$--$16/20$ in Inception. These comparisons differ in both embedding space and sample size, so they do not isolate the effect of $n/d$. The coverage collapse and the centered real pair are rejected in every run in all four spaces.

\textbf{A.11 Coverage losses with small mean displacement.}

The mean-matched mode drop of \S\ref{sec:defaults} removes the location component by subtracting it, which is an intervention on the statistic rather than on the model. We therefore built coverage losses that preserve the sample mean exactly or change it only weakly, and ran the same battery on all five spaces. Two constructions matter. An intra-mode collapse shrinks every point toward its own cluster centroid, preserving the overall mean to numerical zero while destroying within-mode spread. A least-displacing mode drop deletes half of whichever mode moves the mean least, giving $\|\Delta\hat\mu\|/r \le 0.016$ against $0.036$--$0.055$ for the largest-mode drop.

Neither reproduces the dissociation, and they fail in opposite directions. The collapse is detected by the median heuristic in $19$--$20/20$ runs and by the window and MMDAgg in $20/20$ in every space, but by KID's cubic kernel in $0$--$2/20$ at strength $0.10$ and $7$--$25/50$ at $0.25$. It is a mean-preserving coverage failure that the cubic kernel does not see and the median does. The least-displacing drop has low rejection rates: $0$--$7/50$ for all four configurations in all five spaces, MMDAgg included. Empirical level on undistorted controls in the same runs is $0$--$5/50$.

Two consequences follow. The claim that narrow bandwidths detect coverage loss that the defaults miss holds for the subtraction-based construction and does not generalize to coverage losses with small mean displacement, and we scope it accordingly. The cubic kernel's weakness holds across constructions: it is the one configuration with low rejection rates on a coverage failure that every other configuration detects.

\textbf{A.12 Real comparisons through the full battery.} Table~\ref{tab:er1} lists every real pair through the full battery, and Appendix~A.14 gives their geometry.

\begin{table}[t]
\caption{Real comparisons through the full battery at $n{=}780$ per side with $20$ draws; cells with mid-range counts were increased to $50$ draws. ``Recentered'' centers each sample separately on its own mean before permutation. This changes the raw higher moments as well as the means, and the recentered counts are descriptive (Section~\ref{sec:protocol}). Controls are independent splits of the corpus. text2midi \citep{bhandari2025text2midi} is sampled on $2{,}000$ held-out MidiCaps captions; FolkRNN is the model of \citet{sturm2016music}. Audio rows: MusicGen \citep{copet2023simple} and Stable Audio \citep{evans2026stableaudio3}, $2{,}000$ ten-second clips each against $2{,}000$ FMA clips, both sides through one provenance pipeline. $\dagger$: for the Aria row \citep{bradshaw2025scaling} the plain arm is $20/20$ at the median, cubic and MMDAgg but $0/20$ at the window bandwidth.}
\label{tab:er1}
\begin{center}\small
\begin{tabular}{lcccc}
\toprule
 & \multicolumn{1}{c}{Plain} & \multicolumn{3}{c}{Recentered} \\
Pair & all four configs & median & cubic & window / MMDAgg \\
\midrule
GPT-2 vs.\ training corpus & $20/20$ & $25/50$ & $0/50$ & $50/50$ \\
FolkRNN vs.\ training corpus & $20/20$ & $20/20$ & $20/20$ & $20/20$ \\
$T{=}0.85$ vs.\ $T{=}1.0$ & $20/20$ & $20/20$ & $20/20$ & $20/20$ \\
$T{=}0.7$ vs.\ $T{=}1.0$ & $20/20$ & $20/20$ & $20/20$ & $20/20$ \\
text2midi vs.\ MidiCaps (its training set) & $20/20$ & $20/20$ & $20/20$ & $20/20$ \\
GPT-2 vs.\ FolkRNN (model vs.\ model) & $20/20$ & $20/20$ & $20/20$ & $20/20$ \\
DDPM vs.\ CIFAR-10 (vision) & $20/20$ & $0/20$ & $0/20$ & $20/20$ \\
MusicGen vs.\ FMA (PANNs) & $20/20$ & $20/20$ & $0/20$ & $20/20$ \\
MusicGen vs.\ FMA (VGGish) & $20/20$ & $20/20$ & $20/20$ & $20/20$ \\
Stable Audio vs.\ FMA (PANNs) & $20/20$ & $20/20$ & $0/20$ & $20/20$ \\
Stable Audio vs.\ FMA (VGGish) & $20/20$ & $20/20$ & $20/20$ & $20/20$ \\
Aria vs.\ its own continuations & $20/20^{\dagger}$ & $0/20$ & $0/20$ & $0/20$ \\
split control & $0$--$1/20$ & \multicolumn{3}{c}{$0/20$} \\
\bottomrule
\end{tabular}\end{center}
\end{table}

\textbf{A.13 Ranking reversal between two audio generators.} The table gives wins out of $20$ subsamples at $n{=}780$ per side against the FMA reference, and the model with the smaller statistic wins. Configurations rank the two models in opposite orders, and the data do not say which order is correct.

\begin{center}\small
\begin{tabular}{lcccccc}
\toprule
 & median & cubic & window $0.05$ & FMD & location & dispersion \\
\midrule
PANNs: MusicGen & 0 & 0 & 10 & 0$^{*}$ & 0 & 0$^{*}$ \\
PANNs: Stable Audio & 20 & 20 & 10 & 20$^{*}$ & 20 & 20$^{*}$ \\
VGGish: MusicGen & 1 & 2 & 20 & 7 & 3 & 8 \\
VGGish: Stable Audio & 19 & 18 & 0 & 13 & 17 & 12 \\
\bottomrule
\end{tabular}\end{center}
$^{*}$ In PANNs ($d{=}2048$) the sample covariances at $n{=}780$ are rank-deficient. The Bures formula needs no inversion and stays defined, but as a raw scalar it carries a large finite-sample bias there, so the reversal rests on the kernel columns. The location column needs no covariance and carries no such caveat.

\textbf{A.14 Geometry of the real pairs.} The location shares of the Fr\'echet decomposition for eight of the pairs of Table~\ref{tab:er1} are: FolkRNN $0.375$, GPT-2 $0.417$, Aria $0.253$, DDPM vs.\ CIFAR-10 $0.278$, GPT-2 vs.\ FolkRNN $0.437$, text2midi $0.667$, $T{=}0.85$ $0.728$, $T{=}0.7$ $0.737$. The pair whose counts change most after centering, GPT-2 against its corpus, is not the one with the lowest location share. Per-pair Fr\'echet terms, including the audio and model-versus-model pairs, are in the per-cell result files. The shares quoted here use the full embedding sets. The screen of Appendix~A.9 uses a $3{,}000$-sample subsample per side, which is why its DDPM value differs. Known-case validation of the spectrum diagnostic: FolkRNN, a model whose low diversity is documented by its authors, yields ratios $0.55$ (eff), $0.63$ ($n_{90}$), $0.59$ (trace), and $0.54$ (Vendi) against its training corpus in CLaMP~2.

\textbf{A.15 Contraction against the bandwidth grid.} The table gives rejections out of $20$ draws at $n{=}780$ per side with $500$ permutations. ``matched'' is the contraction whose 2-Wasserstein distance equals that of the $s{=}0.1$ mean shift, which bisection places at $s{=}0.100$ in every space, as the Gaussian expression for the two distortions predicts.

\begin{center}\small\setlength{\tabcolsep}{4pt}
\begin{tabular}{llccccccc}
\toprule
Space & Distortion & median & $0.02$ & $0.05$ & $0.1$ & $0.2$ & $0.5$ & cubic \\
\midrule
PANNs & control & 0/20 & 1/20 & 0/20 & 0/20 & 3/20 & 2/20 & 0/20 \\
 & contraction $0.1$ & 20/20 & 19/20 & 20/20 & 20/20 & 20/20 & 20/20 & 0/20 \\
 & contraction, matched & 20/20 & 19/20 & 20/20 & 20/20 & 20/20 & 20/20 & 4/20 \\
 & mean shift $0.1$ & 20/20 & 17/20 & 20/20 & 20/20 & 20/20 & 20/20 & 20/20 \\
\midrule
Inception & control & 0/20 & 1/20 & 2/20 & 2/20 & 0/20 & 0/20 & 1/20 \\
(CIFAR-10) & contraction $0.1$ & 20/20 & 13/20 & 20/20 & 20/20 & 20/20 & 20/20 & 4/20 \\
 & contraction, matched & 20/20 & 10/20 & 20/20 & 20/20 & 20/20 & 20/20 & 4/20 \\
 & mean shift $0.1$ & 20/20 & 3/20 & 20/20 & 20/20 & 20/20 & 20/20 & 20/20 \\
\midrule
Inception & control & 1/20 & 2/20 & 2/20 & 1/20 & 0/20 & 1/20 & 2/20 \\
(STL-10) & contraction $0.1$ & 20/20 & 7/20 & 20/20 & 20/20 & 20/20 & 20/20 & 5/20 \\
 & contraction, matched & 20/20 & 7/20 & 20/20 & 20/20 & 20/20 & 20/20 & 6/20 \\
 & mean shift $0.1$ & 20/20 & 2/20 & 18/20 & 20/20 & 20/20 & 20/20 & 20/20 \\
\bottomrule
\end{tabular}\end{center}
CLaMP~2 and VGGish give the same pattern, and their per-cell counts are in the result files. Strengths are matched by 2-Wasserstein distance throughout, and no other matching criterion was tried. This table is an independent run of the contraction cells, with its own seeds. Where the earlier battery of Table~\ref{tab:a2other} covers the same cell the two differ by at most three draws (cubic contraction: CLaMP~2 $3/20$ here against $4/20$ there, VGGish $3/20$ against $2/20$, PANNs $0/20$ against $2/20$). Table~\ref{tab:main} and the main text quote the counts of this run.

\textbf{A.16 Precision, recall, density and coverage on the same battery.} The table gives rejections out of $20$ draws against split-split nulls with $k{=}5$, $200$ permutations, and $n{=}780$ per side.

\begin{center}\small\setlength{\tabcolsep}{5pt}
\begin{tabular}{lccccc}
\toprule
Distortion & CLaMP~2 & VGGish & PANNs & Inception & STL-10 \\
\midrule
control & 1/20 & 2/20 & 2/20 & 1/20 & 2/20 \\
rotation $s{=}0.1$ & 17/20 & 14/20 & 16/20 & 7/20 & 1/20 \\
mean shift $s{=}0.1$ & 10/20 & 9/20 & 16/20 & 9/20 & 7/20 \\
contraction $s{=}0.1$ & 20/20 & 20/20 & 20/20 & 20/20 & 20/20 \\
mode drop, mean-matched & 6/20 & 1/20 & 3/20 & 4/20 & 3/20 \\
mode drop, least-displacing & 2/20 & 0/20 & 1/20 & 1/20 & 0/20 \\
intra-mode collapse & 20/20 & 20/20 & 20/20 & 20/20 & 20/20 \\
rank-2 rotation & 10/20 & 7/20 & 6/20 & 5/20 & 2/20 \\
rank-1 inflation & 20/20 & 11/20 & 13/20 & 11/20 & 6/20 \\
real pair, recentered & 20/20 & 20/20 & 20/20 & 14/20 & n/a \\
\bottomrule
\end{tabular}\end{center}
The collapse leaves a signature the kernel statistics do not report: recall falls from $0.85$ to $0.13$ and density rises to $10.1$ in Inception. The mean-matched and least-displacing mode drops leave all four quantities within $0.05$ of their control values in every space.

\textbf{A.17 Two properties of the embedding spaces.} The table holds two unrelated blocks with different rows. The left block, over model and reference pairs, gives the spectrum of $\Delta\Sigma$ for the real pairs, as the fraction of $\|\Delta\Sigma\|_F^2$ in the five leading eigendirections and the effective rank of its squared spectrum. The right block, over embedding spaces, gives the mean absolute value of $u = x^\top y / d$, the argument of the cubic kernel. $^{*}$CLIP, the embedding CMMD uses, is included for reference. No result in the paper is computed in that space.

\begin{center}\small
\begin{tabular}{lccc@{\hskip 2em}lcc}
\toprule
Pair & top-5 & eff.\ rank & & Space & $\mathbb{E}|u|$ & $\mathrm{Var}(u)$ \\
\midrule
DDPM vs.\ CIFAR-10 & 0.40 & 59.4 & & CLaMP~2 & 0.504 & 0.0037 \\
GPT-2 vs.\ corpus & 0.65 & 15.0 & & VGGish & $1.98\!\times\!10^{4}$ & $1.16\!\times\!10^{6}$ \\
FolkRNN vs.\ corpus & 0.87 & 6.3 & & PANNs & 0.040 & $4\!\times\!10^{-5}$ \\
text2midi vs.\ MidiCaps & 0.60 & 18.2 & & Inception (CIFAR-10) & 0.144 & 0.0011 \\
MusicGen vs.\ FMA & 0.71 & 12.6 & & Inception (STL-10) & 0.069 & 0.0005 \\
Stable Audio vs.\ FMA & 0.72 & 11.9 & & CLIP (CIFAR-10)$^{*}$ & 0.404 & 0.0035 \\
Aria vs.\ continuations & 0.65 & 17.0 & & & & \\
\bottomrule
\end{tabular}\end{center}
VGGish is the outlier on the right: its features are large enough that the constant in $(u+1)^3$ is negligible, so the cubic kernel there is effectively $u^3$. Raw KID values in that space are therefore numerically extreme and are not comparable with KID values in the other spaces.

\textbf{A.18 Power against sample size in vision.} Inception embeddings of CIFAR-10 are tested against a pool of $100{,}000$ DDPM samples. Cells give rejections out of $20$ draws at $n\le5{,}000$, $10$ at $n{=}10{,}000$, and $5$ at $n{=}20{,}000$. ``rotation'' is matched to the mean shift in 2-Wasserstein distance, and ``real pair'' is DDPM against CIFAR-10 after separate centering (descriptive).

\begin{center}\small\setlength{\tabcolsep}{5pt}
\begin{tabular}{llccccc}
\toprule
Arm & Configuration & $780$ & $2{,}500$ & $5{,}000$ & $10{,}000$ & $20{,}000$ \\
\midrule
rotation & median & 3/20 & 3/20 & 17/20 & 10/10 & 5/5 \\
 & cubic & 2/20 & 0/20 & 1/20 & 2/10 & 2/5 \\
 & window $0.05$ & 5/20 & 20/20 & 20/20 & 10/10 & 5/5 \\
 & MMDAgg & 20/20 & 20/20 & 20/20 & 10/10 & n/a \\
\midrule
real pair & median & 0/20 & 2/20 & 20/20 & 10/10 & 5/5 \\
 & cubic & 0/20 & 0/20 & 0/20 & 0/10 & 2/5 \\
 & window $0.05$ & 20/20 & 20/20 & 20/20 & 10/10 & 5/5 \\
 & MMDAgg & 20/20 & 20/20 & 20/20 & 10/10 & 5/5 \\
\midrule
control & median & 0/20 & 2/20 & 0/20 & 1/10 & n/a \\
 & cubic & 0/20 & 1/20 & 0/20 & 0/10 & n/a \\
 & window $0.05$ & 1/20 & 1/20 & 0/20 & 1/10 & n/a \\
\bottomrule
\end{tabular}\end{center}
Cells marked n/a were not run. The control arm was not extended beyond $n{=}10{,}000$, and the MMDAgg rotation cell was not run at $n{=}20{,}000$. No result is inferred for those cells. A mean shift of $s{=}0.1$ is detected in every configuration at every size.

\textbf{A.19 Injected copies.}
\begin{table}[h]
\caption{Cells give the mean response to duplicate injection in control-normalized units ($z$ against independent splits; $20$ draws per cell, exact copies; perturbed copies agree within $\pm0.5$) and MMDAgg rejections ($5/360$ pooled across all cells and both variants, ${\approx}1.4\%$). Copying lowers cross-set distances, so responsive statistics go negative; the text discusses magnitudes $|z|$. Narrow is the Gaussian kernel at $0.05\times$ median $d^2$, Median the median heuristic, and FMD the Fr\'echet distance on embeddings.}
\label{tab:copying}
\begin{center}
\begin{tabular}{llrrrr}
\toprule
Space & Copies & Narrow & Median & FMD & MMDAgg \\
\midrule
VGGish & $1\%$ & $-0.7$ & $+0.1$ & $-0.4$ & $2/20$ \\
       & $5\%$ & $-3.8$ & $-0.7$ & $-0.8$ & $0/20$ \\
       & $10\%$ & $-8.0$ & $-0.9$ & $-1.6$ & $0/20$ \\
PANNs  & $1\%$ & $-0.6$ & $-0.2$ & $-0.5$ & $0/20$ \\
       & $5\%$ & $-3.3$ & $-0.5$ & $-1.0$ & $0/20$ \\
       & $10\%$ & $-6.8$ & $-0.9$ & $-1.9$ & $0/20$ \\
CLaMP~2 (dedup.) & $1\%$ & $-0.9$ & $-0.3$ & $-0.3$ & $0/20$ \\
       & $5\%$ & $-3.4$ & $-0.3$ & $-1.4$ & $1/20$ \\
       & $10\%$ & $-6.3$ & $-0.7$ & $-2.7$ & $0/20$ \\
\bottomrule
\end{tabular}
\end{center}
\end{table}

Table~\ref{tab:copying} injects training-set copies into one side at $1$, $5$ and $10\%$. Injecting copies from the reference introduces dependence between the two evaluation arms, so this experiment is analysed separately from the independent-sample tests, and the usual calibration guarantee does not apply to it. The narrow-bandwidth statistic, scored against independent splits, responds monotonically and identically for exact and lightly perturbed copies. The median bandwidth changes little. Under this injection MMDAgg rejects in $5$ of $360$ runs. The narrow-bandwidth response is a sensitivity to duplication rather than a copying test, and a copying test needs its own reference set, statistic and calibration.

\textbf{A.20 Spectral diversity across a temperature sweep.}
\begin{table}[h]
\caption{Temperature dose-response of a GPT-2-class model on the folk corpus. Cells are ratios to the training reference, means $\pm$ s.d.\ over five random $5{,}000$-sample subsets of each set.  $\mathrm{eff}$ is the exponentiated spectral entropy of the covariance, $n_{90}$ the number of components covering $90\%$ of the variance, $\mathrm{tr}$ the trace, and Vendi the score of \citet{friedman2023vendi} with a cosine kernel.}
\label{tab:e3}
\begin{center}
\begin{tabular}{lrrrr}
\toprule
$T$ & eff ratio & $n_{90}$ ratio & tr ratio & Vendi ratio \\
\midrule
$0.50$ & $0.79\pm.01$ & $0.80\pm.01$ & $0.84\pm.00$ & $0.75\pm.01$ \\
$0.70$ & $0.88\pm.00$ & $0.87\pm.01$ & $0.95\pm.01$ & $0.83\pm.01$ \\
$0.85$ & $0.92\pm.01$ & $0.89\pm.01$ & $1.02\pm.01$ & $0.90\pm.01$ \\
$1.00$ & $1.08\pm.01$ & $1.07\pm.01$ & $1.00\pm.01$ & $0.99\pm.02$ \\
$1.20$ & $1.04\pm.01$ & $0.97\pm.01$ & $1.21\pm.00$ & $1.15\pm.01$ \\
\bottomrule
\end{tabular}
\end{center}
\end{table}

We sample a GPT-2-class symbolic-music transformer \citep{skiers2024symbotunes} (mini-GPT-2 configuration, trained on the reference corpus) at temperatures $0.5$ to $1.2$ ($7{,}379$ to $10{,}205$ surviving pieces per setting, Appendix~A.4) and embed the outputs with CLaMP~2. Table~\ref{tab:e3} reports summaries normalized by the training reference. Effective dimensionality, $n_{90}$ and the Vendi score rise with temperature (Spearman $+0.90$, $+0.90$ and $+1.00$ over five settings), with an inversion between $T{=}1.0$ and $T{=}1.2$ for the first two. The covariance trace is not monotonic. Trace and spectral shape therefore summarize different aspects of the change, and the trace already falls at the lowest temperature. The effective dimensionality, $n_{90}$ and trace summaries depend only on the covariance spectrum, so they do not respond to a rotation, which leaves the spectrum unchanged. We call them spectral diversity diagnostics: they describe concentration of the spectrum, not coverage in general. Temperature changes several properties at once, so the sweep is a descriptive application. The covariance summaries cost a $d\times d$ eigendecomposition, against an $n\times n$ kernel spectrum for Vendi.

\textbf{A.21 Sensitivity to split-based normalization.} A common calibration scores a statistic in standard deviations of a null estimated from random splits of the corpus, and background-versus-evaluation splits in FAD pipelines invite it. We deduplicated the symbolic corpus (removing the ${\sim}3\%$ of points whose nearest-neighbour squared distance lies below $0.1\times$ the median squared distance) and ran a fixed battery with split-based $z$-scores and with permutation tests. Permutation power did not change beyond sampling variation (rotation at $s{=}0.05$: $3/20$ on the original corpus and $6/20$ on the deduplicated one; at $s{=}0.1$: $20/20$ on both). The standard deviations of split-based nulls at narrow bandwidths changed by up to a factor of $58$, which rescales every $z$-score in the battery by the same factor. Split-based $z$-scores are therefore not comparable across corpora with different duplicate structure. Two readings of split-based $z$-scores do not survive permutation tests. Raw audio embeddings appear to admit no rotation-sensitive bandwidth, while permutation power at $0.05\times$ the median against the rotation at $s{=}0.1$ is $16/20$ in VGGish and $19/20$ in PANNs on the transition-preprocessed embeddings ($20/20$ for both at $s{=}0.5$). Near-duplicates appear to carry the rotation-sensitive bandwidths, while permutation power does not change with deduplication.

\end{document}